\documentclass{article}

    \PassOptionsToPackage{numbers, compress}{natbib}

\usepackage{graphicx}
\usepackage{wrapfig}

\usepackage[preprint]{neurips_2024}

\usepackage[utf8]{inputenc} 
\usepackage[T1]{fontenc}    
\usepackage{hyperref}       
\usepackage{url}            
\usepackage{booktabs}       
\usepackage{amsfonts}       
\usepackage{nicefrac}       
\usepackage{microtype}      
\usepackage{xcolor}         

\newcommand{\Tref}[1]{Table~\ref{#1}}

\newcommand{\Fref}[1]{Figure~\ref{#1}}
\newcommand{\Sref}[1]{Section~\ref{#1}}

\newcommand{\eref}[1]{Eq.~(\ref{#1})}
\newcommand{\fref}[1]{Fig.~\ref{#1}}

\usepackage{xspace}
\usepackage{amsmath}
\usepackage{bbding}

\def\eg{\emph{e.g}.} 
\def\ie{\emph{i.e}.} 
\def\etc{\emph{etc}.~}

\newcommand{\cL}{\mathcal{L}}

\newcommand{\papername}[0]{DreamPolish\xspace}

\title{\papername: Domain Score Distillation With Progressive Geometry Generation}

\author{%
  Yean Cheng$^{*}$\\
  Zhipu AI\\
  \texttt{yean.cheng@zhipuai.cn} \\
  \And
  Ziqi Cai\thanks{Equal contribution.}\\
  Peking University\\
  \texttt{zqtsai@gmail.com} \\
  \And
  Ming Ding\\
  Zhipu AI\\
  \texttt{ming.ding@zhipuai.cn}
  \And
  Wendi Zheng\\
  Tsinghua University\\
  \texttt{zhengwd23@mails.tsinghua.edu.cn}
  \And
  Shiyu Huang\\
  Zhipu AI\\
  \texttt{shiyu.huang@zhipuai.cn}
  \And
  Yuxiao Dong\\
  Tsinghua University\\
  \texttt{yuxiaod@tsinghua.edu.cn}
  \And
  Jie Tang\\
  Tsinghua University\\
  \texttt{jietang@tsinghua.edu.cn}
  \And
  Boxin Shi\thanks{Corresponding author.}\\
  Peking University\\
  \texttt{shiboxin@pku.edu.cn}\\
}

\begin{document}

\maketitle

\begin{abstract}
We introduce \papername, a text-to-3D generation model that excels in producing refined geometry and high-quality textures. In the geometry construction phase, our approach leverages multiple neural representations to enhance the stability of the synthesis process. Instead of relying solely on a view-conditioned diffusion prior in the novel sampled views, which often leads to undesired artifacts in the geometric surface, we incorporate an additional normal estimator to polish the geometry details, conditioned on viewpoints with varying field-of-views. We propose to add a surface polishing stage with only a few training steps, which can effectively refine the artifacts attributed to limited guidance from previous stages and produce 3D objects with more desirable geometry. The key topic of texture generation using pretrained text-to-image models is to find a suitable domain in the vast latent distribution of these models that contains photorealistic and consistent renderings. In the texture generation phase, we introduce a novel score distillation objective, namely domain score distillation (DSD), to guide neural representations toward such a domain. We draw inspiration from the classifier-free guidance (CFG) in text-conditioned image generation tasks and show that CFG and variational distribution guidance represent distinct aspects in gradient guidance and are both imperative domains for the enhancement of texture quality. Extensive experiments show our proposed model can produce 3D assets with polished surfaces and photorealistic textures, outperforming existing state-of-the-art methods. 



\end{abstract}

\begin{figure}
    \centering
    \includegraphics[width=1\linewidth]{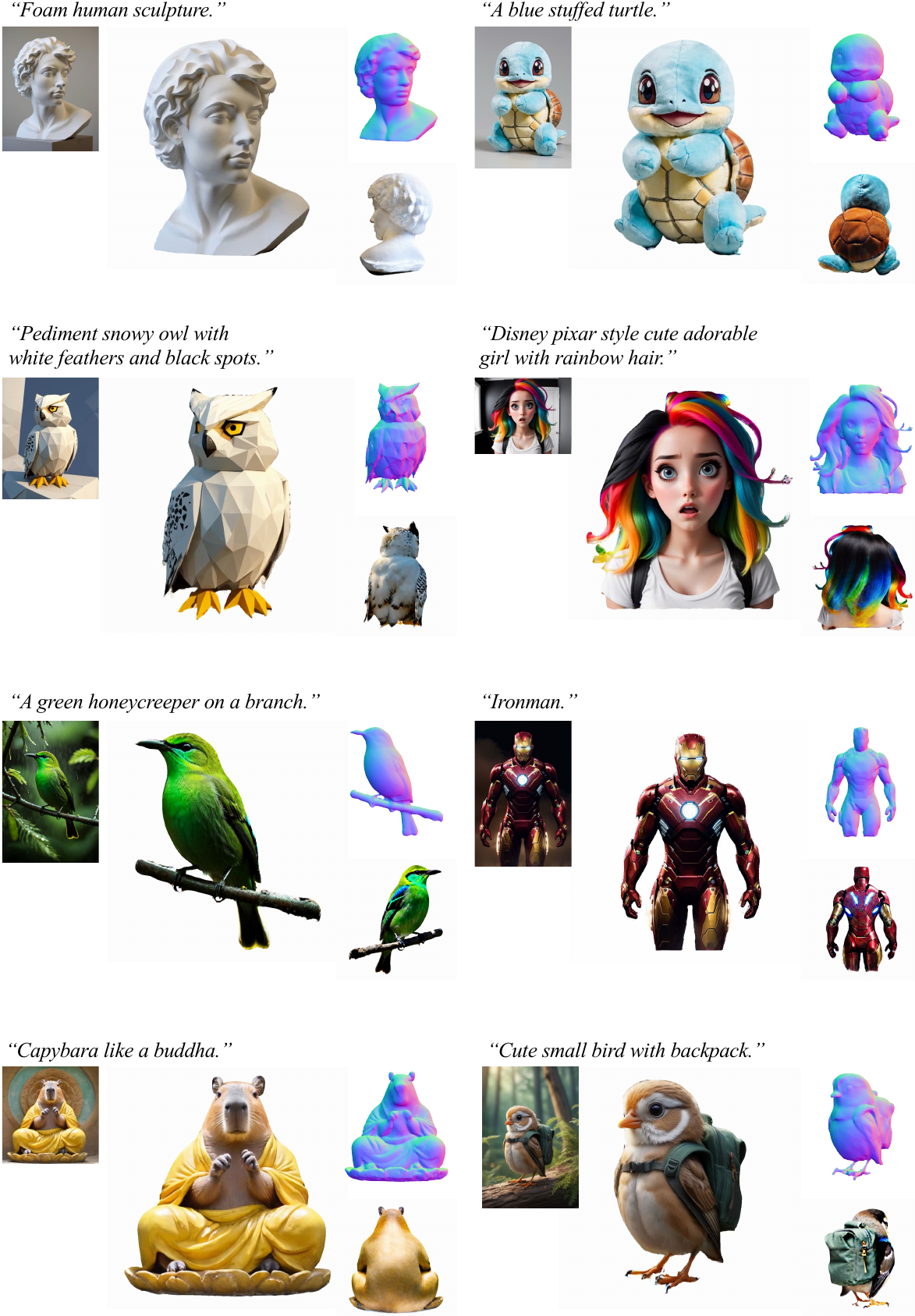}
    \caption{\papername excels in producing 3D models with polished geometry and photorealistic texture. Please refer to the supplemental material for more results and videos.}
    \label{fig:teaser}
\end{figure}

\section{Introduction}
\label{sec:introduction}

Generating 3D assets benefits a series of downstream applications like
virtual reality, movies, video game, 3D printing, \etc Existing 3D generation methods can be divided into 3D native method and 2D/3D hybrid methods. 3D native generation methods~\cite{liu2023zero1to3, sargent2023zeronvs, liu2023syncdreamer} directly model 3D data, and demonstrate considerable success in producing basic 3D objects. However, these methods often struggle to produce complex objects with intricate geometry details and photorealistic textures, primarily due to a shortage of high-quality training data. Recent advances in text-to-image diffusion models~\cite{ho2020denoising, rombach2022high, zhang2023adding} demonstrate the effectiveness of large-scale pretraining on high-quality image-text pairs. The 2D/3D hybrid models, which combine such 2D diffusion models with 3D neural representations, have shown promising outcomes. 


The Score Distillation Sampling (SDS) objective introduced in DreamFusion
~\cite{poole2023dreamfusion} is widely used in the 2D/3D hybrid models. It serves as a bridge to bring photorealistic generation capability embedded in 2D text-to-image models to 3D content generation. During training, renderings with random viewpoints are sampled and sent to a pretrained 2D text-to-image diffusion model (\eg,\cite{rombach2022high,DeepFloydIF,podell2023sdxl}). The pretrained 2D diffusion model is incorporated to leverage the difference in distribution between the renderings and images posteriorly generated by 2D text-to-image models. 


However, using SDS as the sole optimization criterion can lead to inconsistent geometry among viewpoints. Furthermore, to maintain training stability, the SDS loss demands an excessively high classifier-free guidance (CFG) weight, leading to overly saturated outcomes. Recent trials~\cite{lin2023magic3d,Qian2023arXiv,sun2023dreamcraft3d} propose to decompose the text-to-3D process into multiple phases (\eg, geometry construction and texture generation). Under such a strategy, researchers could leverage varying priors on decomposed sub-tasks, hence boosting the generation performance. Despite these advancements, the quality of generated 3D content has not reached the level of handcrafted and professionally produced 3D assets. In the geometric construction phase, the utilization of a view-conditioned diffusion prior leads to artifacts on the generated surfaces, mainly because such a model cannot to offer adequate supervision on geometric intricacies such as surface details. In the texture generation phase, the quality of the resulting texture continues to be inferior compared to those produced by text-to-image diffusion models. Moreover, obtaining higher-quality textures often requires compromising on stability, presenting the challenge of balancing texture photorealism with training stability.



We introduce \textit{\papername}, a text-to-3D generation method that produces 3D objects with polished geometry and photorealistic textures. We first focus on the geometric surface refinement. We employ a variety of neural implicit and explicit representations to progressively retrieve geometry details, and we utilize an additional polishing stage that boosts the surface smoothness with the help of a pretrained normal estimation prior. We further propose a score distillation objective, dubbed domain score distillation (DSD), designed to identify the target domain that can ensure stability throughout the distillation process and enhance the photorealism of textures. 
Our pipeline consists of two phases: \textit{progressive geometry polishing} and \textit{domain-guided texture enhancing}.



In the progressive geometry polishing phase, we start with a text prompt and its corresponding image, and construct a neural representation with image-level supervision in the reference view and employ vanilla SDS from a view-conditioned diffusion prior~\cite{liu2023zero1to3} in novel views. We use NeRF~\cite{MildenhallSTBRN20} in the early iterations and progressively change the neural representation to advanced surface representations (\ie, NeuS~\cite{wang2021neus} and DMTet~\cite{shen2021dmtet}). However, the sampled field-of-view remains constrained by the pretrained distribution of the view-conditioned diffusion prior, and there is no surface supervision in novel views. To further polish the geometry surface, we incorporate a diffusion prior that produces 3D-consistent normal estimation~\cite{fu2024geowizard} to refine and polish the surface in the novel views. The initial limitation to the field-of-view, dictated by the pretrained distribution of 3D diffusion priors, is eventually relaxed, enabling the normal estimator to address and rectify any preceding artifacts. 



In the domain-guided texture enhancing phase, we first demonstrate through empirical analysis why the vanilla SDS tends to be unstable, requiring a disproportionately high weight for CFG. Subsequently, we illustrate why recent works~\cite{wang2023prolificdreamer,sun2023dreamcraft3d} using guidance from variational domains improve the texture quality and effectively reduce the CFG weight to its normal ranges. In the conditional text-to-image generation, the use of classifier-free guidance leveraging unconditional image domains is shown to mitigate the decline in generated performance observed when text conditioning is introduced. This suggests that the unconditional image domain is beneficial for generation diversity and stability. Building upon these analyses, we propose DSD, aiming to balance the generation quality and stability. DSD mimics the classifier-free guidance generation process and proposes to use the unconditional gradient direction to guarantee the stability of score distillation.

As illustrated in \Fref{fig:teaser}, \papername demonstrates its capability in generating 3D content featuring polished geometry and photorealistic textures. We have conducted extensive experiments to showcase the superior performance of \papername compared to previous methods. Additionally, we have performed ablation studies to validate the effectiveness of our proposed modules. The complete implementation will be made publicly available upon acceptance of our work.


\section{Related work}
\label{sec:related work}


\textbf{3D representations} are the foundation for accurate and detailed text-to-3D generation. These representations can be categorized into explicit, implicit, and hybrid forms. Explicit representations, such as point clouds, voxel grids, and meshes, provide precise control by directly describing 3D geometry. However, integrating these forms into deep learning frameworks is challenging due to their lack of differentiability~\cite{li2024advances,Zhan2021MultimodalIS}. Implicit representations, including NeRF~\cite{MildenhallSTBRN20}, NeuS~\cite{wang2021neus}, Gaussian splatting~\cite{kerbl3Dgaussians} and Instant-NGP~\cite{nerf2022ngp}, leverage neural networks to generate continuous volumetric fields, making them more suitable for deep learning applications. Implicit representations produce fine-grained 3D models, however, they pose challenges in terms of editability and controllability~\cite{Gao2022NeRFNR}, and require additional tools like MarchingCubes~\cite{lorensen1998marching} for rasterization. A hybrid approach such as DMTet~\cite{shen2021dmtet}, Plenoxels~\cite{yu2022plenoxels} and TensoRF~\cite{Chen2022ECCV} combines the advantages of both explicit and implicit methods. DMTet~\cite{shen2021dmtet} utilizes a deformable tetrahedral grid to optimize surface geometry and topology, converting signed distance functions into explicit meshes for high-resolution 3D synthesis. We propose using a series of representations to progressively generate 3D assets, leveraging the advantages of different forms of representations.

\textbf{3D diffusion priors} extend the capabilities of diffusion models to 3D data, improving the quality of generation and reconstruction. Notable techniques such as the Zero123 series (\eg, Zero123~\cite{liu2023zero1to3} and Zero123++~\cite{Shi2023Zero123AS}) and GeoWizard~\cite{fu2024geowizard} incorporate geometric priors to ensure the consistency and coherence in 3D structures. GeoWizard~\cite{fu2024geowizard} combines depth and normal information with a U-Net~\cite{Ronneberger2015UNetCN}, and excels in providing high-quality surface normal estimations from single images, leveraging large-scale 3D datasets like OmniData~\cite{eftekhar2021omnidata} to mitigate data limitations. These methods show significant potential in enhancing 3D generation quality. We use a combination of 3D diffusion priors to eliminate the ambiguity in novel views due to lack of information, thereby improving the robustness and photorealism of 3D asset generation.

\textbf{Text-to-3D generative models} have been extensively studied to automate the creation of 3D assets, reducing the need for manual intervention. These models can be classified into native 3D models and 2D/3D hybrid models. Native 3D models such as ZeroNVS~\cite{sargent2023zeronvs}, Text2Shape~\cite{chen2019text2shape}, ShapeCrafter~\cite{fu2022shapecrafter}, SyncDreamer~\cite{liu2023syncdreamer} and DreamGaussian~\cite{Tang2023dreamgaussian}, generate 3D data directly using specialized neural networks. These models employ complex neural architectures to create detailed 3D shapes conditioned on text and viewpoints. However, these 3D native models fall short when generating intricate objects attributing to limited representation ability. In contrast, 2D/3D hybrid models, such as DreamFusion~\cite{poole2023dreamfusion}, Magic3D~\cite{lin2023magic3d}, Fantasia3D~\cite{Chen2023ICCV}, DreamControl~\cite{huang2023dreamcontrol}, and HIFA~\cite{zhu2023hifa} combine pretrained 2D diffusion models with 3D consistency constraints. Hybrid models harness the diverse generation capabilities of 2D models while ensuring geometric coherence in 3D space, addressing the challenges of data scarcity and high computational demands inherent to native 3D models. Despite these advances, the quality of generated 3D content is still inferior to that of the handcrafted 3D content in terms of surface quality and texture photorealism. Our method addresses these limitations, producing both high-quality surfaces and realistic textures.

\section{Preliminaries}
Diffusion models \cite{sohl2015deep, ho2020denoising} are a collection of likelihood-based generative models used to learn data distributions. Forward process is a Gaussian 
process that progressively adds 
random noise \(\epsilon\) to the data \(\mathbf{x}\) sampled from \( q(\mathbf{x})\). 
The reverse process reconstructs the original data from \(\mathbf{x}_T\). The generative process is parameterized as a Markov process: \(
p_\phi(\mathbf{x}_{0:T}) := p(\mathbf{x}_T) \prod_{t=1}^{T} p_\phi(\mathbf{x}_{t-1}|\mathbf{x}_t).
\)
The diffusion model is trained to maximize the variational lower bound of the data log-likelihood~\cite{ho2020denoising}. 
The objective has been optimized into the mean squared error form by the researchers through a series of mathematical derivations and the re-parameterization trick for better convergence:
\begin{equation}
    \mathcal{L}_{\text{Diffusion }}(\phi):=\mathbb{E}_{\mathbf{x} \sim q\left(\mathbf{x}\right), t \sim \mathcal{U}(0,1), \epsilon \sim \mathcal{N}(\mathbf{0}, \mathbf{I})}\left[\omega(t)\left\|\epsilon_\phi\left(\alpha_t \mathbf{x}+\sigma_t \epsilon; t\right)-\epsilon\right\|_2^2\right].
    \label{eq:obj}
\end{equation}
Later research attempt to add text condition on \(y\) to the generation process~\cite{rombach2022high, balaji2022ediffi}, which can be expressed as \( \epsilon_\phi\left(\mathbf{x}_t; y,t\right)\). The diversity of the generated images and the alignment with the text conditions are two variables that need to be traded off. The CFG weight \cite{ho2022classifier} is employed to resolve the balance between these two aspects. CFG employs the implicit approach to train diffusion models under conditional and unconditional modes, enabling the estimation of \(\nabla_{\mathbf{x}_t} \log q\left(\mathbf{x}_t|y\right)\) and \(\nabla_{\mathbf{x}_t} \log q\left(\mathbf{x}_t\right)\). During the sampling process, the strength of the diversity and alignment can be parameterized as a weighted form of conditioned and un-conditioned objectives,
\begin{equation}
    {\epsilon}_{\phi}(\mathbf{x}_t ; y,t)\rightarrow \epsilon_{\phi}(\mathbf{x}_t;y,t )+\omega\cdot\left[\epsilon_{\phi}(\mathbf{x}_t ; y,t)-\epsilon_{\phi}(\mathbf{x}_t;t )\right],
    \label{eq:cfgcfg}
\end{equation}
where \(\omega\) is the guiding scale controlling the trade-off between fidelity and diversity. In a typical image generation task, generally setting \(\omega \in [7.5, 12.5]\) can yield the best generation results.

Dreamfusion~\cite{poole2023dreamfusion} proposes SDS, effectively integrating the diffusion models into the text-to-3D generation. SDS uses a pretrained diffusion model \(\epsilon_\phi\) to emulate the conditioned posterior distribution of the 3D model generation, \ie, for each image rendered from the differentiable 3D representation \(\theta\), SDS uses the diffusion model to verify if the generated image aligns with the distribution of images generated by the diffusion model \(p_\phi(\mathbf{x}_t | y)\). This optimization is interpreted as minimizing the following KL divergences for all \(t\), and can be simplified as,
\begin{equation}
\nabla_\theta \mathcal{L}_{\text{SDS}} = \mathbb{E}_{t, \epsilon, \mathbf{c}} \left[ w(t) (\epsilon_{\phi}(\mathbf{x}_t ;y,t) - \epsilon) 
\frac{\partial \mathbf{x}}{\partial \theta} \right],
\label{eq:sds}
\end{equation}
where \(\theta\) refers to the learnable parameters of a certain differentiable 3D representation (\eg, NeRF \cite{MildenhallSTBRN20}, DMTet \cite{shen2021dmtet}).
However, balancing consistency and photorealism is challenging. For example, DreamFusion~\cite{poole2023dreamfusion} requires a significantly high 
\(\omega=100\) to achieve a consistent 3D model, which results in unrealistic artifacts such as over-smoothed texture and abnormally high saturation~\cite{wang2023prolificdreamer, sun2023dreamcraft3d}. Recently, DreamCraft3D~\cite{sun2023dreamcraft3d} proposes a bootstrapped score distillation (BSD) paradigm to improve the texture generation quality. BSD builds on VSD~\cite{wang2023prolificdreamer} and proposes to utilize a low-rank adaptation (LoRA) of the pretrained diffusion prior to modeling the variational distribution of each 3D object, and use generated images at different views to finetune the original model (\ie, a Dreambooth~\cite{ruiz2023dreambooth}) to further ``bootstrap'' the performance:
\begin{equation}
\nabla_\theta \mathcal{L}_{\text{BSD}} = \mathbb{E}_{t, \epsilon, \mathbf{c}} \left[ w(t) (\epsilon_{\text{DreamBooth}}(\mathbf{x}_t ;y,t) - \epsilon_{\text{LoRA}}(\mathbf{x}_t ;y,t)) 
\frac{\partial \mathbf{x}}{\partial \theta} \right].
\label{eq:vsd}
\end{equation}
\vspace{-1.5em}
\section{Method}
\label{sec:method}
\begin{figure}[t]
    \centering
    \includegraphics[width=0.99\linewidth]{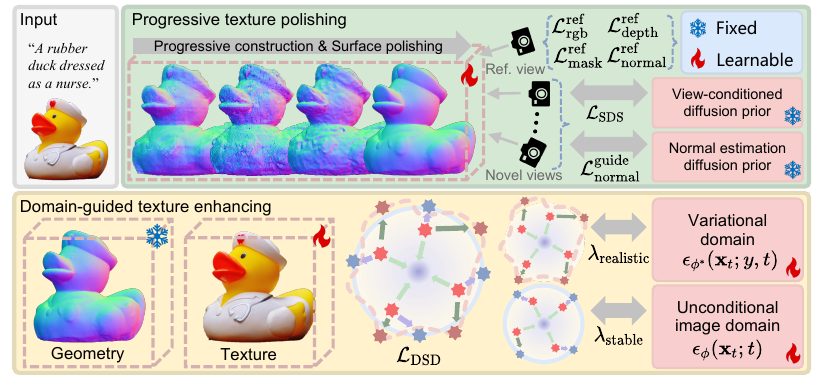}
    \caption{Overview of \papername. Given a text prompt and its corresponding generated image shown in the top left as input, \papername first progressively constructs a fine-grained 3D geometry 
    with
    coherent and smoothed surface. Then, \papername leverages DSD as the score distillation objective to guide the representation towards a domain with both consistent and photorealistic texture.}
    \vspace{-1em}
    \label{fig:pipeline}
\end{figure}

The overall pipeline is illustrated in \Fref{fig:pipeline}. We decompose the intricate text-to-3D task into two phases: progressive geometry polishing and domain-guided texture enhancing. In this section, we first describe the geometry construction process detailed in \Sref{subsec:Geometry construction}. Next, we introduce the texture generation method in \Sref{subsec:Texture generation}. 


\subsection{Progressive geometry polishing}
\label{subsec:Geometry construction}

\paragraph{Progressive construction.} 
Explicit and implicit neural representations exhibit both strengths and limitations. We propose to leverage the strengths of different representations by progressively generating 3D content with a combination of these 3D representations. The generation starts with NeRF~\cite{MildenhallSTBRN20} due to its stable training process and ability to quickly generate a rough 3D structure. We then transition to NeuS~\cite{wang2021neus}, a surface-based representation that provides more accurate and detailed surface information. Inspired by \cite{li2023neuralangelo}, we incorporate a progressive
hash band to further enhance the quality of NeuS~\cite{wang2021neus}. Subsequently, we switch to DMTet~\cite{shen2021dmtet}, which allows for the use of NVDiffrast~\cite{Laine2020diffrast} and integration of the graphics pipeline, enabling fast rendering.

The sampled views can be divided into a reference view (the view with input image as a reference) and novel views (novel synthesized views without any reference). We employ different training objectives on these views, dubbed as reference-based loss \(\mathcal{L}^{\text{ref}}\) and guidance-based loss \(\mathcal{L}^{\text{guide}}\). The reference-based loss \(\mathcal{L}^{\text{ref}}\) primarily measures the error between the rendered image under the reference frame and the reference image itself. \(\mathcal{L}^{\text{ref}}\) can be considered a reconstruction loss:
\begin{equation}
    \mathcal{L}^{\text{ref}} = \lambda_1 \cL^{\text{ref}}_{\text{normal}} + 
    \lambda_2 \cL^{\text{ref}}_{\text{depth}} + \lambda_3 \cL^{\text{ref}}_{\text{mask}} + \lambda_4 \cL^{\text{ref}}_{\text{rgb}}. 
\end{equation}
where \(\lambda_{*}\) are the respective loss weights, and \(\{\cL^{\text{ref}}_{\text{normal}}, \cL^{\text{ref}}_{\text{depth}}, \cL^{\text{ref}}_{\text{mask}}, \cL^{\text{ref}}_{\text{rgb}}\}\) are the losses for surface normal, depth, foreground mask, and reference image, respectively, defined as follows:
\begin{align}
    \cL^{\text{ref}}_{\text{rgb}} =||\hat{\mathbf{m}}\odot(\hat{\mathbf{x}} - g(\theta, \hat{\mathbf{c}}))||_2, & \quad \cL^{\text{ref}}_{\text{mask}} =||\hat{\mathbf{m}} - g(\theta; \hat{\mathbf{c}})||_2,
\end{align}
where \(\hat{\mathbf{c}}\) is the camera pose corresponding to the reference image \(\hat{\mathbf{x}}\). We add normal and depth supervision based on prediction from an off-the-shelf normal and depth estimator~\cite{eftekhar2021omnidata}:
\begin{align}
    \cL^{\text{ref}}_{\text{normal}} = - \frac{\mathbf{n}\hat{\mathbf{n}}}{||\mathbf{n}||_2||\hat{\mathbf{n}}||_2}, & \quad \cL^{\text{ref}}_{\text{depth}} = - \frac{\text{conv}(\mathbf{d}, \hat{\mathbf{d}})}{\sigma(\mathbf{d})\sigma(\hat{\mathbf{d}})}.
\end{align}
The reference-based loss function above ensures that the 3D model generates high-quality images at pose \(\hat{\mathbf{c}}\). However, there is no corresponding reference pose under other camera coordinates, the model needs to rely on pretrained models (\ie, priors) to determine geometry information in those novel poses. Although view-conditioned native 3D models have inferior rendering performance compared to 2D/3D models, they contain 3D coherent information and can be directly used for multi-view geometric supervision. For novel views, we use the vanilla SDS Loss, as in  \Fref{eq:sds}, for supervision:
\begin{equation}
    \nabla_\theta \mathcal{L}^{\text{guide}} = \nabla_\theta \mathcal{L}_{\text{SDS}}(\phi_1, g(\theta)). 
\end{equation}
where \(\phi_1\) is the Stable-Zero123 \cite{liu2023zero1to3} model. Combining the above operations, the gradient of the geometry generation loss can be expressed as:
\begin{equation}
    \nabla_\theta \mathcal{L}^{\text{geom}} = \lambda_{\text{guide}}\nabla_\theta\mathcal{L^{\text{guide}}} + \lambda_{\text{ref}}\nabla_\theta\mathcal{L^{\text{ref}}}. 
\end{equation}
where \(\lambda_{\text{ref}}\) and \(\lambda_{\text{guide}}\) are the corresponding model weights.

\paragraph{Surface polishing.} Despite the strengths of our progressive construction pipeline, we observe that the generated geometry could still benefit from additional refinement. We introduce a surface polishing stage aiming at further enhancing the surface. In this stage, we fix the texture parameters and solely update the geometry to polish the geometry. Additionally, we loosen the field-of-view camera restrictions of Stable Zero123~\cite{liu2023zero1to3}, allowing for a greater diversity of camera parameters. This enables rendering novel views with more diverse viewing conditions, which in turn provides richer information for the polishing process.

We utilize a pretrained diffusion model~\cite{fu2024geowizard} to predict normal maps from images rendered under novel views, eliminating the ambiguity of geometry caused by insufficient information, we leverage standard SDS loss as supervision:
\begin{equation}
\mathcal{L}_{\text{normal}}^{\text{guide}}= \lambda_{\text{comp}}||\hat{\mathbf{n}}_{\text{comp}}-\hat{\mathbf{n}}_{\text{novel}}||_2+\lambda_{\text{lpips}}\mathcal{L}_{\text{lpips}}(\hat{\mathbf{n}}_{\text{comp}},\hat{\mathbf{n}}_{\text{novel}}),
\end{equation}
where $\hat{\mathbf{n}}_{\text{comp}}$ is rendered normal map, $\hat{\mathbf{n}}_{\text{novel}}$ is corresponding normal prediction generated from rendered RGB image, and $\mathcal{L_\text{lpips}}$ denotes a perceptual loss~\cite{DBLP:conf/cvpr/ZhangIESW18}.

\subsection{Domain-guided texture enhancing}
\label{subsec:Texture generation}

\begin{figure}[t]
    \centering
    \includegraphics[width=0.99\linewidth]{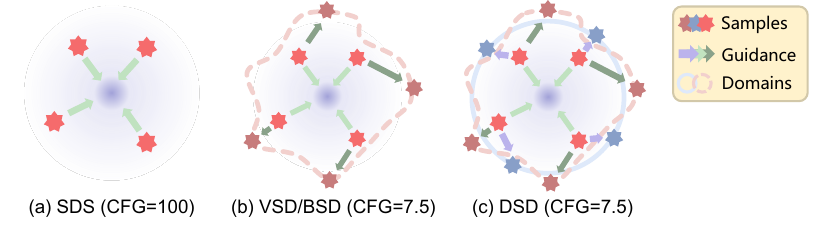}
    \caption{Illustration on different score distillation strategies. (a): Vanilla SDS~\cite{poole2023dreamfusion} only has guidance direction on zero-mean noise; (b): VSD~\cite{wang2023prolificdreamer} and BSD~\cite{sun2023dreamcraft3d}  utilize a variational domain to improve texture quality; (c): our proposed DSD provides guidance directions toward unconditional image domain and variational domain, further improving the stability and photorealism of rendered texture.}
    \vspace{-1em}
    \label{fig:dsd}
\end{figure}

With the geometry construction stage, our pipeline yields 3D objects represented by DMTet~\cite{shen2021dmtet} with detailed geometric intricacies and smoothed surfaces. In the following phase, we fix the geometry representation and focus on texture generation.

We illustrate the score distillation process in the latent space as \fref{fig:dsd}. Diffusion models are designed to predict noises, which are assumed to follow a \(D\) dimensional Gaussian distribution. The major population of the probability is  concentrated on a ring with a radius 
\(\sqrt{D}\) around the origin, visualized as the solid circle in \fref{fig:dsd}. Samples around this ring represent the photorealistic images learned by pretrained text-to-image models from large scale image datasets.

The standard SDS objective (\eref{eq:sds}) minimizes distance between the estimated noise \(\epsilon_{\phi}(\mathbf{x}_t ;y,t)\) from diffusion model and random noise \(\epsilon\). However, \eref{eq:sds} is a sum of the expected distance between the estimated noise and a random sampled noise. Due to its zero mean nature, the \(\epsilon\) term represents a force toward the center (mean) of the distribution, as demonstrated in \fref{fig:dsd} (a), which eventually causes unsatisfactory over-saturated results. The recently proposed VSD~\cite{wang2023prolificdreamer} and BSD~\cite{sun2023dreamcraft3d} provide guidance direction towards a variational domain, represented by the dotted line in \fref{fig:dsd} (b). This domain is the learned distribution of the object being constructed, hence providing high-quality knowledge to distill from compared to the vanilla SDS term. However, this variational domain is parameterized by a LoRA~\cite{hu2021lora}, the training process does not guarantee the learned domain is stable and contains the necessary information for high-quality texture with sufficient diversities. The unconditional image domain is located around the major distribution, and represents the diversity and stability of the text-to-image generation capability~\cite{ho2022classifier}. 

We further propose DSD, which balances quality and stability within the distillation sampling process. As visualized in \fref{fig:dsd} (c), DSD consists of two guidance from the variational domain and the unconditional image domain, we employ the variational domain guidance to maintain photorealism and utilize the unconditional image guidance to ensure the stability of the distillation process. The gradient of the proposed DSD distillation sampling method can be expressed as:
\begin{equation}
\nabla_\theta \mathcal{L}_{\text{DSD}} = \mathbb{E}_{t, \epsilon, \mathbf{c}} \left[ w(t) (\epsilon_{\phi}(\mathbf{x}_t ;y,t) - \lambda_{\text{realistic}}\epsilon_{\phi^*}(\mathbf{x}_t ;y,t) - \lambda_{\text{stable}}\epsilon_{\phi}(\mathbf{x}_t;t))
\frac{\partial \mathbf{x}}{\partial \theta} \right],
\label{eq:dsd}
\end{equation}
where \(\lambda_{\text{realistic}}\) and \(\lambda_{\text{stable}}\) represents the weights of diversity and stability.

\section{Experiment}

\label{sec:experiments}
\subsection{Experimental settings}
\paragraph{Baselines.} To evaluate the performance of our method, we compare it with several representative and state-of-the-art models. Specifically, we compare our method to DreamFusion~\cite{poole2023dreamfusion}, GeoDream~\cite{Ma2023GeoDream}, and ProlificDreamer~\cite{wang2023prolificdreamer}, which use text inputs only, as well as Magic123~\cite{Qian2023arXiv} and DreamCraft3D~\cite{sun2023dreamcraft3d}, which support both text and reference image inputs. 

\paragraph{Datasets.} We use a diverse dataset composed of text-image pairs for a comprehensive evaluation. Additionally, we generate normal maps and depth maps using Omnidata~\cite{eftekhar2021omnidata}, following the methodology of DreamFusion~\cite{poole2023dreamfusion}. 

\paragraph{Error metrics.} To assess the quality of the generated results, we calculate PSNR~\cite{huynh2008scope}, SSIM~\cite{DBLP:journals/tip/WangBSS04}, and LPIPS~\cite{DBLP:conf/cvpr/ZhangIESW18} in the reference view. Following the methodology in \cite{sun2023dreamcraft3d}, we also employ CLIP Score~\cite{hessel2021clipscore} to measure the consistency between the generated model and the text prompt. 

\subsection{Qualitative comparison}
\begin{figure}[t]
    \centering
    \includegraphics[width=0.99\linewidth]{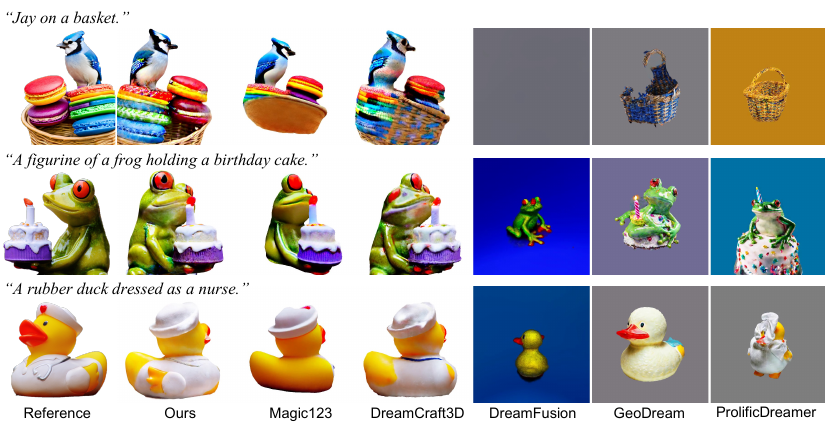}
    \caption{Qualitative comparisons with baseline methods. Our method produces 3D objects with high-quality geometry and photorealistic textures. Please refer to the supplementary for more results.}
    \label{fig:comp}
\end{figure}

We compare our method with the aforementioned baselines. The results are demonstrated in \fref{fig:comp}. Given the reference image, we show the generated renderings of our model and the baselines under novel viewpoints. Benefiting from refined geometry and the DSD objective, our model could generate intricate details and photorealistic textures, such as the frog's eye in the second row and the wing of the rubber duck in the third row. Please refer to the supplement for more results.

\begin{table}[t]
\centering
\caption{Quantitative Comparison Results. $\uparrow$ ($\downarrow$) indicates that higher (lower) is better. We highlight the best score in boldface.}
\label{tab:quantitative evaluation}
\begin{tabular}{lllll}
\toprule
\textbf{Model}           & \textbf{PSNR $\uparrow$} & \textbf{SSIM $\uparrow$} & \textbf{LPIPS $\downarrow$} & \textbf{CLIP Score $\uparrow$} \\ \hline
Magic123~\cite{Qian2023arXiv}            & 20.30           & 0.803           & 0.148            & 0.720                \\ 
DreamCraft3D~\cite{sun2023dreamcraft3d}       & 24.40           & \textbf{0.933}         & 0.093   & 0.754                \\ 
Ours         & \textbf{25.13}    &  \textbf{0.933} &  \textbf{0.087}  &  \textbf{0.759}     \\ \bottomrule
\end{tabular}
\vspace{-1em}
\end{table}

\subsection{Quantitative experiments}
\paragraph{Comparision results.} We render images from novel views and compute the mean CLIP Score between each pair of rendered images and the text prompts. \Tref{tab:quantitative evaluation} summarizes the results of the quantitative evaluation. Our method outperforms the baselines in every metric.
\begin{wrapfigure}{r}{0.4\textwidth}
    \centering
    \includegraphics[width=1\linewidth]{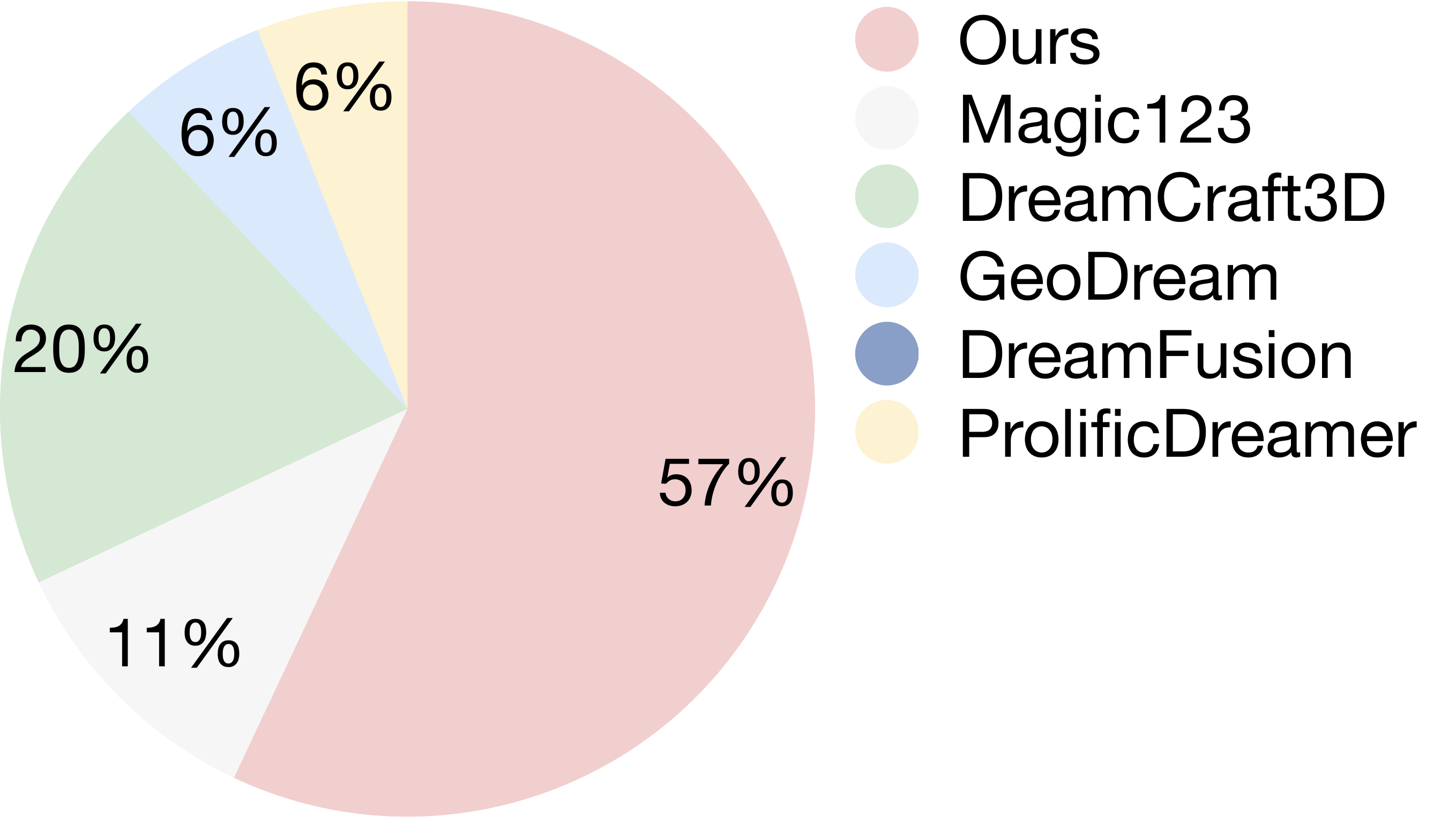}
    \vspace{-3mm}
    \caption{User study results.}
    \label{fig:user study}
    \vspace{-15mm}
\end{wrapfigure}
\paragraph{User study.} We randomly select multiple examples from different viewpoints and shuffle the data from our model and the baseline methods. Users are asked to choose the model they consider to have the best performance for each example. We received a total of 40 valid responses, and the results are shown in \Fref{fig:user study}. It is evident that our method outperforms the other methods and is well-received by the general users.

\subsection{Ablation study}
\begin{figure}[t]
    \centering
    \includegraphics[width=0.95\linewidth]{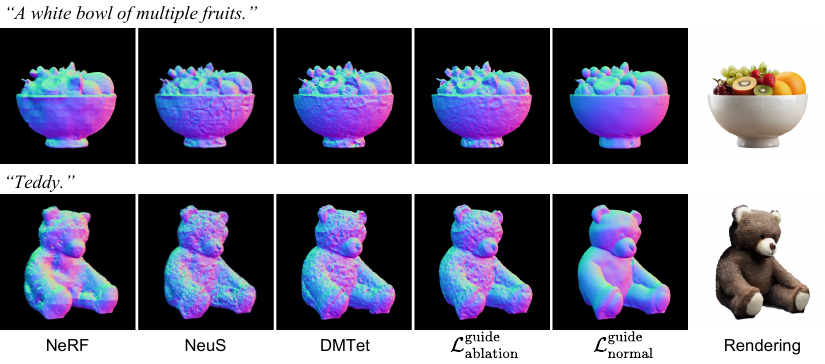}
    \caption{Ablation study on geometry construction. Geometric quality and surface smoothness in varying representations are progressively refined along the training process. In the surface polishing stage, normal smooth loss \(\mathcal{L}^{\text{novel}}_{\text{ablation}}\) is insufficient for surface smoothing while the proposed \(\mathcal{L}^{\text{novel}}_{\text{normal}}\) objective can effectively polish the artifacts generated in previous stages.}
    \vspace{-1em}
    \label{fig:ab-geom}
\end{figure}

\paragraph{Geometry construction.} \Fref{fig:ab-geom} shows the normal map of each stage in the geometry construction process. We replace the proposed $\mathcal{L}^{\text{guide}}_{\text{normal}}$ with a simple normal smooth loss:  $\mathcal{L}^{\text{guide}}_{\text{ablation}} = |\nabla_h \mathbf{n}|_2^2 + |\nabla_w \mathbf{n}|_2^2$, where $\nabla_h \mathbf{n}$ and $\nabla_w \mathbf{n}$ represent the gradients of the normal map. \(\mathcal{L}^{\text{guide}}_{\text{ablation}}\) polishes the artifacts in previous stages into refined surface. As shown in the fourth row, the smooth loss \(\mathcal{L}^{\text{guide}}_{\text{ablation}}\) still produces artifacts and is insufficient for surface polishing.

\begin{figure}[t]
    \centering
    \includegraphics[width=0.99\linewidth]{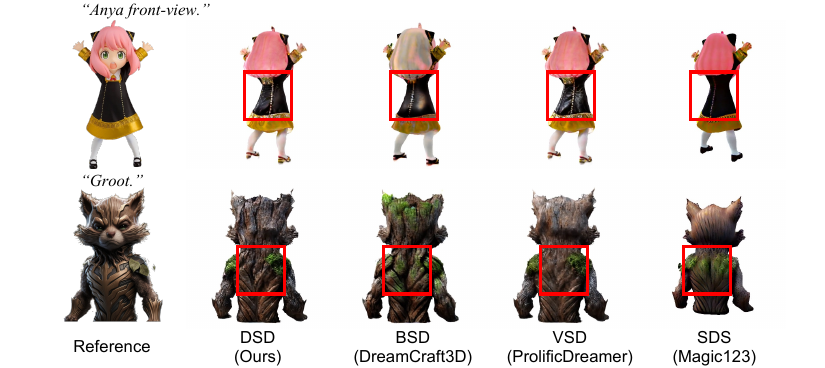}
    \caption{Ablation study on texture generation. With the same fixed geometry, the proposed DSD objective produces textures with the most photorealistic details.}
    \vspace{-1em}
    \label{fig:ab-texture}
\end{figure}

\paragraph{Texture generation.} \Fref{fig:comp} proves our model generates state-of-the-art 3D objects, thanks to both the geometry construction and texture generation modules. To further prove the effectiveness of the DSD objective, we use the previous losses (SDS~\cite{poole2023dreamfusion}, VDS~\cite{wang2023prolificdreamer} and BSD~\cite{sun2023dreamcraft3d}) on the same geometry produced by our geometry phase, and compare the texture quality with that produced by our proposed DSD. The results are shown in \fref{fig:ab-texture}, demonstrating that the DSD can generate textures with superior photorealism 
(\eg, fewer artifacts in the first row and more details in the second row).
\vspace{-1em}

\section{Conclusion}
We present \papername, a text-to-3D generation model that achieves polished geometry and photorealistic textures. By progressively constructing geometry and incorporating a surface polishing stage, \papername could produce refined surfaces; with the proposed DSD objective, our approach addresses key challenges in texture generation and is able to stably distill photorealistic details from the latent domain. Extensive experiments demonstrate that \papername can produce 3D assets with superior quality, setting a new benchmark in the 3D generation field.

\paragraph{Limitations.} 
\label{subsec:limitations}
Despite the success of \papername, several limitations remain. The geometry refinement stage, while effective, is limited by the quality of the initial geometry. Additionally, the computational cost of our approach can be further optimized.





\newpage

{\small
\bibliographystyle{plain}
\bibliography{main}
}








\end{document}